\documentclass{article}
\usepackage{ijcai18}

\usepackage{times}
\usepackage{xcolor}
\usepackage{soul}
\usepackage[utf8]{inputenc}
\usepackage[small]{caption}

\usepackage{graphicx}
\usepackage{subcaption}
\usepackage{color}
\usepackage{amsmath}

\graphicspath{{imgs/}}

\DeclareMathOperator*{\argminA}{arg\,min}

\title{Action Categorization for Computationally Improved Task Learning and Planning}
\author{
Lakshmi Nair \and 
Sonia Chernova
\\ 
Georgia Institute of Technology \\
lnair3@gatech.edu,
chernova@cc.gatech.edu
}

\begin{document}
\maketitle

\begin{abstract}
This paper explores the problem of task learning and planning, contributing the \textit{Action-Category Representation (ACR)} to improve computational performance of both Planning and Reinforcement Learning (RL). ACR is an algorithm-agnostic, abstract data representation that maps objects to action categories (groups of actions), inspired by the psychological concept of \textit{action codes}. We validate our approach in StarCraft and Lightworld domains; our results demonstrate several benefits of ACR relating to improved computational performance of planning and RL, by reducing the action space for the agent. 
\end{abstract}

\section{Introduction}
Research in Psychology has shown that humans handle the complexity of the real world by biasing or constraining their action choice at a given moment based on known object-related actions.  In particular, recent fMRI studies show that the human brain uses \textit{action codes} -- automatically evoked memories of prototypical actions that are related to a given object -- to bias or constrain expectation on upcoming manipulations \cite{schubotz2014objects}.  In effect, given an object, our brain simplifies the action selection process by constraining the decision to a predefined set of known actions.  For instance, a knife and an apple seen together evoke the action codes of ``cutting apple with knife'' and ``peeling apple with knife''. 

Our work presents an analogous mechanism for computational agents, showing that automatically generated action groupings can be used to improve the computational efficiency of both task planning and learning by constraining the action space. We present the \textit{Action Category Representation (ACR)}:  an algorithm-agnostic, abstract data representation that encodes a mapping from objects to action categories (groups of actions) for a task.  Specifically, we incorporate the idea of \textit{action codes} as the action categorization mechanism. We formally define an action code as the tuple:

\centerline{$((o_1, o_2 ... o_j),(a_1, a_2 ... a_k))$}

\noindent Where $(o_1, ... o_j)$ represents a set of objects and $(a_1, ... a_k)$ represents the set of actions associated with them for the task. For instance, the action code corresponding to the knife and apple example above is $((apple,knife), (peel,cut))$. In our work, we use action codes to build the Action Category Representation that can be used to improve computational performance in both task planning and reinforcement learning.

Action codes are closely related to the concept of object affordances \cite{gibson1977perceiving,mcgrenere2000affordances}, which are defined as action possibilities available to the agent for a given object.  Affordances function by priming specific actions for the user by virtue of the object's physical properties (shape, size etc.). In contrast, action codes do not derive from the physical properties of objects, rather from the associative memories of what we use the objects for during everyday tasks. Thus, the notion of affordances is often independent of the task \cite{ellis2000micro,tucker1998relations} while action codes take the task into account.  ACR builds on the notion of action codes, enabling an agent to learn object-action mappings based on prior experience.  

\begin{figure}[t]
\centering
\includegraphics[width=0.35\textwidth]{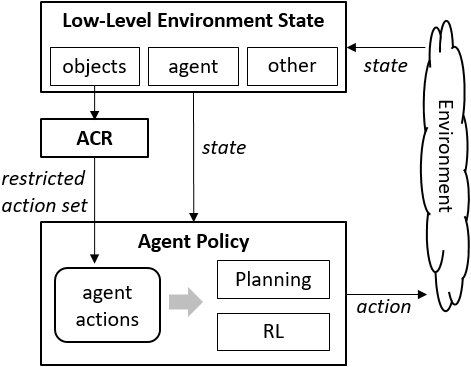}
\captionsetup{width=\linewidth}
\caption{Objects in the low-level environment state are mapped via ACR to action categories to restrict the action set used in the planning or RL techniques}
\label{fig:ACR_flow}
\end{figure}

Within a computational framework, the primary benefit of ACR is to reduce the choice of actions the agent must consider.  Thus, ACR serves as a layer of abstraction between low-level state information and the learning or planning technique used to control the agent (Figure \ref{fig:ACR_flow}). In this paper we:

\begin{enumerate}

\item describe the process of constructing ACR from the agent's experience or human demonstration of a task;

\item show that ACR has formal computational bounds that guarantee its use leads to at the worst case the same, and in the common case much improved, computational performance over traditional techniques that consider objects and actions without categorization;

\item present the computational benefits of using ACR in conjunction with PDDL planning to reduce planning time; and

\item present the computational benefits of using ACR with Q-learning to achieve improved learning performance.

\end{enumerate}

We validate ACR performance in two virtual domains: StarCraft and Lightworld.  We conclude the paper with a discussion of our work and potential future uses of ACR.

\section{Related Work}
In this section, we position our paper in relation to existing work. 

\subsection{Affordance Learning}
As discussed above, the concept of action codes is closely related to action affordances and affordance learning.  Affordances model relationships between individual object \textit{properties} (shape, size, color etc.) to actions and observed effects and are formally defined as 

\noindent \textit{(effect,(object,behavior))} \cite{csahin2007afford}. In contrast, action codes and by extension ACR, relate only semantic labeling and a holistic perception of objects such as ``cup'' or ``box'' to appropriate actions for a task. 

Traditional approaches to affordance learning often involves ``behavioral babbling'' \cite{stoytchev2005behavior,montesano2008learning,lopes2007affordance} wherein the agent physically interacts with objects in a goal-free manner to discover their affordances. Hence, the resulting affordance representation is dissociated from a task, focusing instead on object properties. Such approaches involve several agent-object interactions affecting the scalability of the learning process, making it unfeasible in situations where there is an implicit cost or time constraint on the robot. ACR helps mitigate this cost by the grouping of actions into categories.  

Two works closest to our approach are \cite{kjellstrom2011visual} and \cite{sun2010learning}.  In \cite{kjellstrom2011visual}, Kjellstrom et al.  describe an approach to visual object-action recognition that use demonstrations to categorize semantically labeled objects based on their functionality. This approach bridges the gap between affordance learning and task context since the learning is coupled with a task demonstration. However, it is unclear how the system would incorporate previously unseen objects unless they are observed from additional demonstrations. For instance, given a demonstration of pouring water into a ``cup'', the agent would require additional demonstrations to identify the similar functionality of a ``bowl''. 

Sun et al. in \cite{sun2010learning} learn visual object categories for affordance prediction (Category-Affordance model), reducing the physical interactions with the objects. They use visual features of objects to categorize them on the basis of their functionality. However, it is unclear how the agent would deal with changing features and categories \cite{min2016affordance}, since the model is learned offline as compared to ACR which allows online learning of new objects and categories (Details in Sec 3). Regardless, their approach highlights some of the benefits of categorization on the scalability of learning, which motivates our work.

\subsection{Precondition Learning}
Preconditions can be expressed using predicates which may or may not relate to object affordances. For instance, ``At'' or ``isEmpty'' are object states whereas ``graspable'' is an affordance predicate \cite{lorken2008grounding}. Object-Action Complexes or OACs \cite{geib2006object} include instances of affordances as preconditions in the OAC instantiation. Their approach learns an ``object'' after physical interaction with it, i.e, there is no notion of an object prior to the interaction. For instance, the representation of a cube is learned after the agent grasps a planar surface. Other approaches such as \cite{ekvall2008robot} learn high-level task constraints and preconditions from demonstrations. In contrast to these approaches, ACR categorizes objects on the basis of action codes to improve planning performance as well as the learning performance of RL algorithms.

\subsection{Learning from Demonstration} 
Human demonstrations have been used for both high-level task learning and low-level skill learning \cite{chernova2014robot}; a traditional assumption of LfD is that the human demonstrator is an expert, and the demonstrations are examples of desirable behavior that the agent should emulate. Our work focuses on high level task learning, but considers demonstrations more broadly as examples of what the agent \textit{can} do, rather than what it \textit{should}. This interpretation of the data enables our technique to benefit even from non-expert human users. Demonstration errors can be classified to one of 3 categories \cite{chernova2014robot}: Correct but suboptimal (contains extra steps), conflicting or inconsistent (user demonstrates 2 different actions from the same state) and entirely wrong (user took a wrong action) and we demonstrate the robustness of ACR to suboptimal demonstrations in Sec 6. 

\subsubsection{LfD in planning} 
Abdo et al. in \cite{abdo2012low} discuss the learning of predicates by analyzing variations in demonstrations. The learned predicates are then applied to plan for tasks and accommodate for environmental changes. Kadir et al. in \cite{uyanik2013learning} demonstrates execution of a task by leveraging human interactions. The agent interacts with all of the objects using \textit{all} of the precoded behaviors in its repertoire and uses forward chaining planning to accomplish the task goal. However, with increasing number of behaviors and objects, the search space for the planner can become quite large. Our approach using ACR can help reduce the action space making planning easier. 

\subsubsection{LfD in RL}
Thomaz and Breazeal in \cite{thomaz2006reinforcement} discuss the effect of human guidance on an RL agent. Similar to our approach with ACR, the teacher guides the action selection process to reduce the action space for the RL agent. While both expert and non-expert guidance improved performance when compared to unguided learning, the final performance was sensitive to the expertise of the teacher.

Another well-known approach to integrating LfD and RL is Human-Agent Transfer or HAT \cite{taylor2011using}. Their approach uses a decision list to summarize the demonstration policy with a set of rules. However, it is sensitive to the number and optimality of the demonstrations \cite{suay2016learning,brys2015reinforcement}. We compare ACR to HAT in Sec 6 to demonstrate the benefits of ACR in terms of quantity and quality of the demonstrations.

\subsection{Object Focused Approaches}
In general, recent work in AI and Robotics has increasingly focused on modeling state not simply as a vector of features, but as a set of objects, such as Object-Oriented MDPs (OO-MDP), leading to improved computational performance due to data abstraction and generalization.  In the context of reinforcement learning (RL), \textit{Object-focused Q learning (Of-Q)} represents the state space as a collection of objects organized into object classes, leading to exponential speed-ups in learning over traditional RL techniques \cite{cobo2013object}. More closer to our work, human input containing object-action associations has been used to effectively guide policy learning in Mario \cite{krening2016object}. The input advice is analogous to action codes, eg. ``Jump over an enemy''. Approaches such as \cite{barth2014affordances,cruz2014improving,wang2013robot} have used affordances in RL to prune the action space and improve learning. However, the formalisms in all these approaches differ from ACR and were not extended beyond Reinforcement Learning to planning of tasks.

\section{Action-Category Representation (ACR)} 

The objective of ACR is to categorize objects based on the action codes of a task.  In this section, we first present an example that illustrates the functionality of ACR and contrasts it with object affordance models.  We then present the ACR formalism.
 
As an example, consider the task of packing a cardboard box during clean-up. The action codes for the cardboard box in this case are \textit{((box),(close))} and \textit{((box),(move))}. Based on these action codes, ACR groups the actions \textit{close} and \textit{move} into a single action category associated with the item \textit{box}.  One of the benefits of ACR is that other objects sharing the same action codes, such as \textit{cooking pot} in a dish-clearing task and \textit{suitcase} in a travel-packing task, also become associated with the same action category, enabling the agent to reason about groups of similar objects across tasks that share action codes, despite the physical dissimilarities of the objects.

In our human example, a person seeing a knife and an apple may be primed to \textit{cut} or \textit{peel}, but may also select to ignore this bias and choose to wash the apple instead.  In the agent's case, we similarly have the choice of treating ACR as a hard constraint on the actions available to the agent or as a flexible bias.  In the sections below, we show how planning is well suited to use ACR as a hard constraint, and how ACR can be naturally combined with RL as a bias.  We discuss possible extensions of this view in the conclusion of the paper.

In this paper, we show how ACR can be utilized in two ways. First, during task planning or learning, ACR improves computational efficiency by pruning the action space. Second, given an object not previously seen by the agent, ACR reduces the number of agent-object interactions required to learn its action associations for the task. Note that, in humans, action codes act as a bias and not a strict restriction on actions.  In other words, a person seeing the knife and apple next to each other is primed to perform the actions \textit{cut} and \textit{peel}, but may override this bias too and \textit{put away} the apple instead. In the agent's case, we have the option to treat ACR as a hard constraint on the actions available to the agent or as a flexible bias that also allows re-expanding the action set. In the sections below, we show how planning is well suited to use ACR as a hard constraint, and how ACR can be naturally combined with RL as a bias.  We discuss possible extensions of this view in the conclusion of the paper.

To construct ACR, the agent requires observations of objects in its environment and what actions are related to each object. These observations can be gained either through the agent's own exploration of the environment, or, more effectively, from a human teacher performing demonstrations of the task. During the observation phase, the agent maintains a log of action codes based on the actions performed and the objects that the actions were executed upon.  We define $O$ as the set of all objects in the task environment, and $A$ as the set of all actions pertaining to the task.  An observation log consists of a set of action codes and is represented by $L = \{\hat{c}_1, \hat{c}_2, ... \hat{c}_n\}$, where each timestep in the log is represented by an action code $\hat{c}_i = ((o_1, o_2 ... o_j),(a_1, a_2 ... a_k))$ with $o_j \in O$ and $a_k \in A$.  

\begin{figure}[t]
\centering
\includegraphics[width=0.4\textwidth]{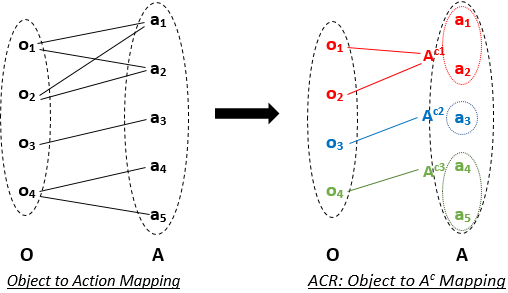}
\captionsetup{width=\linewidth}
\caption{Bipartite graphs representing the relationship from objects to actions (left), and objects to action categories (right)}
\label{fig:OAR_graph}
\end{figure}

The act of building object-action relations can be formulated as a bipartite graph partitioning problem involving the action set $A$ and the objects set $O$. Given a graph $G(V,E)$, with vertices $V$ and edges $E$, the graph is \textit{bipartite} when the vertices can be separated into two sets, such that $V = A \cup O$,  $A \cap O = \emptyset$, and each edge in $E$ has one endpoint in $A$ and one endpoint in $O$. In the context of ACR, $A$ represents actions, $O$ represents objects and an edge $\{a_i, o_i\}$ exists if action $a_i \in A$ and object $o_i \in O$ co-occur within any action code $\hat{c}_k \in L$. For instance, the action code \textit{((box), (push))} is represented by an edge from the action \textit{push} to the object \textit{box}. The resulting bipartite graph has a many-to-many association between objects and actions (Figure \ref{fig:OAR_graph} left).  In ACR, the bipartite graph is generated incrementally from the action codes in the observation log.

The main computational units of ACR are \textit{action categories}, defined by a group or set of actions $A^c \subseteq A$.  Given the bipartite graph above, for a given action $a_j \in A$, let $\hat{O}_{a_j}$ represent the set of objects for which that action co-occurs in some action code (i.e. the edge $\{a_j, o_k \in \hat{O}_{a_j}\}$ exists).   Then we define an action category $A^c$ as:

\centerline{$A^c = \{a_j \ : \bigcup \hat{O}_{a_j} = \bigcap \hat{O}_{a_j}\}$}

This is interpreted as, ``The set of all actions $a_j$ such that union over all $\hat{O}_{a_j}$ is equal to intersection over all $\hat{O}_{a_j}$''. In other words, the action category $A^c$ contains a set of actions that are associated to the same set of objects, allowing us to group all those actions as one set.  If we consider action categories as vertices themselves, then what results is a reduced one-to-many bipartite graph between action categories $A^c$ and the set of objects $O$ (Figure \ref{fig:OAR_graph} right), which is the representation we refer to as \textit{ACR}. Note that the entire set of actions can be grouped into categories such that $A = A^{c_1} \cup A^{c_2} \cup A^{c_3} \cup ... A^{c_n}$.  We define $C = \{A^{c_1}, A^{c_2} ... A^{c_n}\}$ as the set of all action categories learned from observations. Note that, as in most prior work, we assume single-parametric actions \footnote{While it is possible to decompose multi-parametric actions to single-parametric actions as described in \cite{bach2014affordance}, we currently do not model them explicitly within ACR.} \cite{montesano2008learning,ugur2011goal,csahin2007afford}, and that preconditions and effects of those actions are known and can be perceived when planning \cite{agostini2015using,ekvall2008robot}.

The construction of ACR is an online process, allowing learning of new objects and action categories over time with changes in the environment or task. As new action codes are learned, objects or actions can be incorporated by adding them to the graph, along with corresponding edges.  A new action category $A^{c_i}$ may be added to $C$ when a new combination of associated actions is discovered, such that $A^{c_i} \neq A^{c_k}$ $\forall A^{c_k} \in C$. The resulting representation provides an automatically-generated online grouping of objects into categories based on action codes. 

In this paper, we discuss characteristics of ACR that contribute to its novelty and significance:
\begin{enumerate}
    \item groups actions based on action codes in order to reduce the action space for the agent, 
    \item contains and appropriately represents algorithm-agnostic information for planning, as well as RL, to improve their computational performance,
    \item minimizes agent-object interactions for learning the action associations of a new object; and 
    \item requires one or few human demonstrations and is robust to the optimality of these demonstrations.
\end{enumerate}

\section{Computational Performance Analysis}
In this section, we present performance guarantees of ACR; in the following section we then validate our findings with case studies in StarCraft and Lightworld \cite{konidaris2007building} domains.

\subsection{Mathematical Analysis}
We define the total number of actions in a domain to be $|A| = n$, allowing us to bound the total possible action categories to be $|C| \leq 2^n - 1$, representing all possible action combinations from $1$ to $n$ actions. Then a given task involves a subset of these action categories $S \subseteq C$ and a set of objects $O$. The agent is assigned the task of learning $S$ and categorizing the objects in $O$ from observations of action codes.

One of the benefits of ACR is seen when the agent encounters a new set of objects $O'$ (not previously seen) and must discover which actions in $A$ are related to each object in $O'$ for the task execution. Below we present performance analysis of ACR and the baseline that uses no action categorization, with respect to the number of agent-object interactions prior to learning all the actions related to an object for the task ($A_{obj}$). Fewer $A_{obj}$ is computationally preferred since this reduces the number of agent-object interactions, making the learning or planning faster.  

\subsubsection{$A_{obj}$ Without Categorization (Baseline)}
Without categorization, each action is considered independently, in which case to determine the set of actions applicable to a new object the agent must test out all $|A| = n$ actions on that object. That is, $A_{obj} = n$.

\subsubsection{$A_{obj}$ With Action Categories (ACR)}
The use of ACR can improve computational performance through action selection, enabling the agent to more effectively identify (or rule out) object interactions. In the presence of action categories our goal is to, with as few actions as possible, identify the category of a newly discovered object. To do so, we select actions from $A$, and for every attempted action that is unassociated (or associated) with the object we eliminate any action category in $S$ with (or without) that action from further testing. We use entropy as a measure of the most informative action to test so as to eliminate as many action categories as possible with each action tested. The entropy of an action $a$ is given by:

\smallskip \centerline{$H(a) = -p \log(p) - (1-p) \log(1-p)$}

\smallskip \centerline{Where, $p = \frac{\sum_{i=1}^{n} |\{a\} \cap A^{c_i}|}{|S|}$} 

\smallskip \noindent The term $p$ denotes the probability that the action categories contain the action $a$ which is used to compute the entropy. Then the action that minimizes entropy is the most informative action. Therefore, the action $\hat{a}$ chosen for testing is given by:

\smallskip \centerline{$\hat{a} = \argminA_{a \in A} H(a)$}

In the best case, the category may be learned with a single action and hence the lower bound on $A_{obj}$ is 1. The worst case upper bound on $A_{obj}$ remains $n$: $1 \leq A_{obj} \leq n$.

That is, with categorization the performance is \textit{never} worse but typically better than without categorization. In practice, $S$ is usually a small subset of $C$, and therefore $A_{obj} << n$. In the case that a new action category must be learned, which occurs rarely in closed world domains such as StarCraft and Lightworld, $A_{obj}$ = n.  In fact, in the experiments described below the agent obtains all possible action codes even from a single demonstration, allowing all the relevant action categories to be known prior to planning and learning.

\section{Case Study in StarCraft}
In this section, we first briefly describe our domain and highlight the complexity of the problem before discussing the computational benefits of using ACR with planning.

\begin{figure}[t]
\centering
\includegraphics[width=0.45\textwidth]{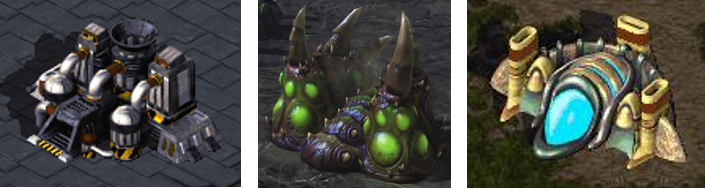}
\captionsetup{width=\linewidth}
\caption{Refineries in StarCraft: Terrans, Zergs and Protoss (left to right), showing their distinct physical appearances}
\label{fig:extractors}
\end{figure}

StarCraft is a real-time strategy game which involves managing one of the 3 diverse civilizations (Terrans, Protoss and Zergs), producing buildings and units while destroying all of your opponents. Across the 3 civilizations, there are over 100 diverse units/buildings and reducing the number of agent-object interactions in this case makes the problem of task planning in StarCraft much more tractable. Figure \ref{fig:extractors} shows a ``refinery'', one of the buildings in StarCraft that is used to extract a gaseous mineral. Given the distinct appearances of the buildings across the 3 civilizations, it may be challenging to identify their related actions by using only physical features without any actual interaction.

\subsection{ACR Extracted from StarCraft}
We extracted ACR from one human demonstration in the Terrans civilization where the teacher successfully completed an in-game mission of creating a defense. Replay logs that summarize the action codes within the demonstration are readily available for StarCraft and are used to construct the ACR. In the case of physical systems, it is possible to extract action codes from human demonstrations using verbal communication during the demonstration or approaches such as \cite{gupta2007objects}.

\begin{table}[ht]
\centering
\includegraphics[width=0.35\textwidth]{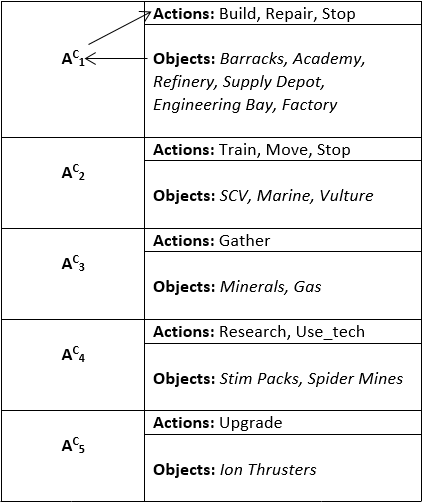}
\captionsetup{width=\linewidth}
\caption{ACR built from human demo}
\label{fig:OAR_int}
\end{table}

Table \ref{fig:OAR_int} shows the complete ACR built from the human demonstration, highlighting the different action categories and object mappings to the action categories. We use this learned representation in the following section to describe its computational benefits.

\subsection{Computational Benefits of ACR with Planning}
We demonstrate two computational benefits of ACR with planning in terms of:

\begin{enumerate}
    \item reduced number of object interactions or $A_{obj}$ required to learn all object-action associations prior to planning in StarCraft (Exploration phase)
    \item improved planning performance due to reduced action space, demonstrated with combat formations in StarCraft
\end{enumerate}

\subsubsection{Benefits of ACR During Exploration Phase}
We demonstrate the benefits of ACR on $A_{obj}$ using build order planning. Build orders dictate the sequence in which units and structures are produced. Prior to planning for a task, there is usually an exploration phase during which the actions associated with the objects (in this case, for the units/structures specified in the build order) are first identified \cite{ugur2011goal}. This exploration phase adds to the overall planning time. Thus, a reduced $A_{obj}$ in the exploration phase would reduce overall planning time. We compare ACR and baseline (without categorization) on $A_{obj}$, during exploration phase of build order planning.

We explore build orders from all 3 civilizations. Table \ref{fig:build_order} shows the total number of agent-object interactions during the exploration phase, along with the number of previously unseen objects for which the object-action relations had to be learned.

\begin{table}[t]
\centering
\includegraphics[width=0.45\textwidth]{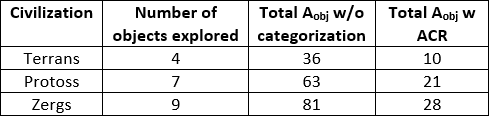}
\captionsetup{justification=centering}
\caption{Total $A_{obj}$ for the different build order exploration phases}
\label{fig:build_order}
\end{table}

As shown in the Table \ref{fig:build_order}, the number of object interactions with ACR is significantly reduced compared to the baseline approach that does not use categorization. In the baseline case, every action (of the 9 actions shown in Table \ref{fig:OAR_int}) has to be attempted on each new object in the build order to discover all of its associations which is mitigated by the use of action categories. The results obtained here highlight the benefits previously discussed in the mathematical analysis of Section 4.1. While the feedback for an invalid action in StarCraft is instantaneous and incurs no significant time cost, in other domains such as task execution with robots, there may be implicit time and cost constraints associated with each interaction. Hence, with ACR it is possible to minimize interactions with the environment by grouping actions.

\subsubsection{Improved Planning Performance with ACR}
In this section, we combine ACR with an existing off-the-shelf PDDL Planner (Fast Forward) \cite{helmert2006fast} to demonstrate how ACR reduces the action space that the planner has to contend with, thus reducing the planning time. 

\begin{figure}[ht]
\centering
\includegraphics[width=0.9\linewidth]{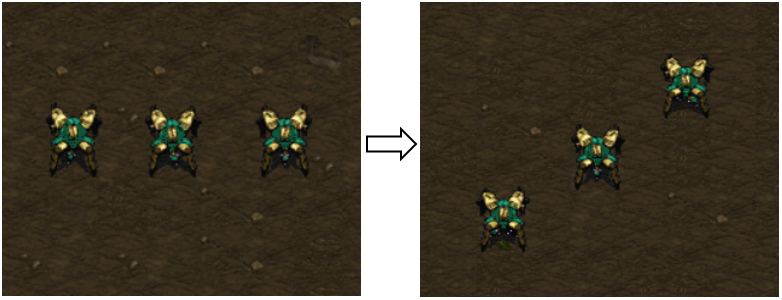}
\captionsetup{width=\linewidth}
\caption{Combat formation problem showing initial (left) and goal (right) states of formation using 3 Dragoons}
\label{fig:planning_problem}
\end{figure}

For this evaluation we use a combat formation problem where combat units (Dragoons) have to form a particular arrangement on a section of the battlefield. We compared the classical planning approach without action categories (baseline) to planning with ACR. We increase the number of Dragoons and demonstrate its effect on the two planning approaches. Figure \ref{fig:planning_problem} shows a sample initial and goal states for the Dragoon formation. 

\begin{figure}[ht]
\centering
\includegraphics[width=\linewidth]{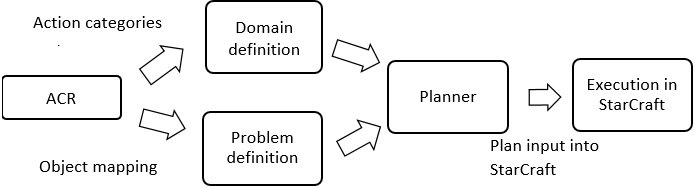}
\captionsetup{width=\linewidth}
\caption{Pipeline for integrating ACR with PDDL planner for planning in StarCraft}
\label{fig:PDDL_pipeline}
\end{figure}

The overall pipeline for combining ACR with the planner is shown in Figure \ref{fig:PDDL_pipeline}. Contrary to the classical approach, ACR introduces the learned action categories into the domain and problem definitions for planning, leading to computational improvements. The classical planning approach instantiates each Dragoon as a separate entity with distinct variables, while the ACR based approach instantiates all the Dragoons in terms of the action category that they are mapped to, which is $A^{c_2}$ in this case (Table \ref{fig:OAR_int}). The domain and problem definitions are thus automatically generated from ACR. The plan is then generated using the domain and problem definitions.

\begin{figure}[tbh]
\centering
\begin{subfigure}[b]{0.4\textwidth}
\includegraphics[width=\linewidth]{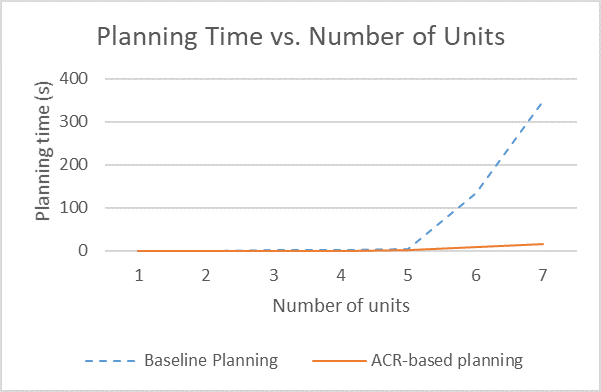}
\caption{Planning time vs. number of Dragoon units}
\end{subfigure}
\hspace{0.25em}
\begin{subfigure}[b]{0.4\textwidth}
\includegraphics[width=\linewidth]{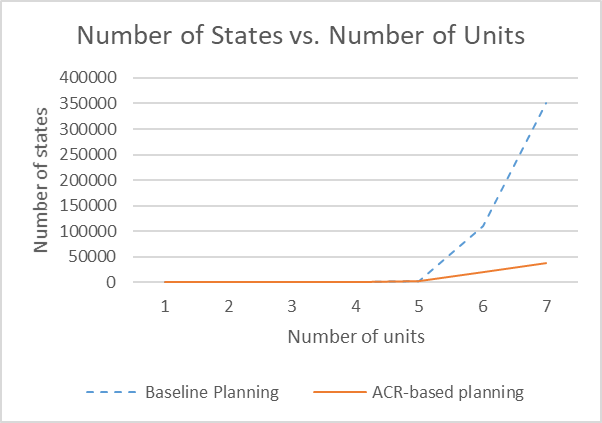}
\caption{Number of search states vs. number of Dragoon units}
\end{subfigure}
\caption{Graphs showing effect of number of Dragoons on planning time (Fig \ref{fig:planning_result}a) and number of search states (Fig \ref{fig:planning_result}b) for the baseline planning and ACR-based planning approaches}
\label{fig:planning_result}
\end{figure}

As shown in Figure \ref{fig:planning_result}, increasing number of Dragoons from 1 to 7, exponentially increases the search time for the classical planning approach as compared to the ACR-based approach (Figure \ref{fig:planning_result}a). This is because the number of states increases exponentially with the number of Dragoons. For instance, on a 5x5 grid, the number of states for 3 Dragoons is ${25 \choose 3}*3 = 6900$ for the classical planning approach and ${25 \choose 3} = 2300$ for the ACR-based planner since ACR instantiates all Dragoons in terms of their action category. Similarly, in the case of forward chaining planners such as \cite{uyanik2013learning}, ACR can help reduce the branching factor from $n_o*|A|$ (where, $n_o$ is the total number of objects or Dragoons in this case, and $|A|$ is total number of actions) to $\sum_{1}^{n} |o^{c_i}|*|A^{c_i}|$ (where, $|o^{c_i}|$ indicates number of objects mapped to action category $A^{c_i}$). Thus, ACR leads to computational benefits by pruning the action space the planner has to contend with. 

\section{Case Study in Lightworld}

In this section, we discuss the computational benefits of applying ACR with RL. We first discuss our domain design, inspired by the Lightworld domain used in the Options RL framework \cite{konidaris2007building}. We then discuss the benefits of applying ACR with RL.

\begin{figure}[t]
\centering
\includegraphics[width=0.3\textwidth]{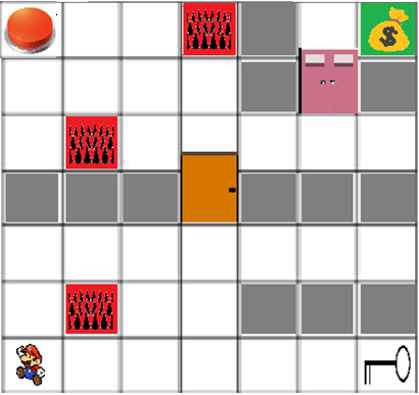}
\captionsetup{justification=centering}
\caption{Example domain}
\label{fig:lightworld}
\end{figure}

Figure \ref{fig:lightworld} shows a sample domain. The game consists of a 7x8 grid of locked rooms with some doors operated by a switch and some doors operated by a key. The goal of the agent is to unlock the doors and move to the final reward. The agent has to move over the button or the key to either press or pick up the object. There are also spike pits that the agent needs to avoid while navigating the room. The agent receives a reward of +100 for reaching the goal state, and a negative reward of -10 for falling into spike pits which are terminal states. Additionally, the agent receives a negative step reward of -0.04. There are a total of 6 actions with each of the four grid directions, a pickup action and press action. The environment is deterministic and unsuccessful actions do not change the state.

\subsection{Integrating ACR with RL}

\begin{figure}[t]
\centering
\includegraphics[width=0.3\textwidth]{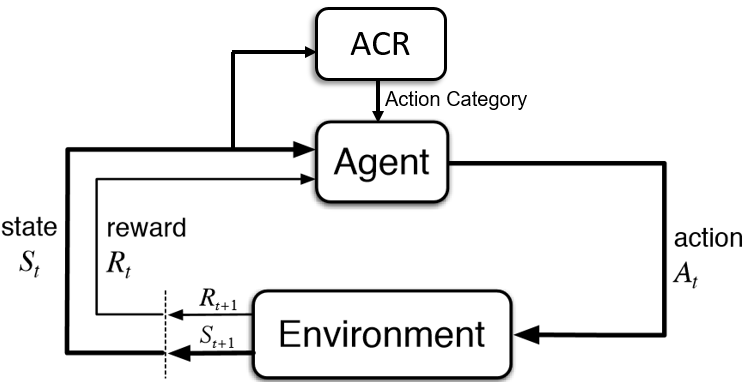}
\captionsetup{justification=centering}
\caption{Pipeline for integrating ACR with RL}
\label{fig:ACR_RL}
\end{figure}

Figure \ref{fig:ACR_RL} shows the integration of ACR with general RL algorithms. ACR influences the action selection step, given an observed state. With ACR, the agent chooses an action from within an action category $A^{c_i} \subset A$ of the known objects it can interact with, given the state. For previously unseen objects whose action categories are unknown, the agent chooses the entropy-wise selected action as described in Sec 4.1 to simultaneously infer the action category of the objects during learning. If the agent cannot interact with any objects, it chooses from the non object-related set of actions (analogous to an ``agent'' action category) given by, $A-\bigcup_{i=1}^{m} A^{c_i}$ where $m$ denotes the number of learned action categories. This reduces the action space for the agent. 

As noted in Sec 3, we use ACR with RL to bias the initial learning rather than applying it as a hard constraint over the entire learning phase. We achieve this by allowing ACR to influence the action choice for a fixed number of episodes, denoted by $N_{ACR}$. This allows the agent to leverage the reduced action space while also learning the optimal policies in states where the ACR-guided policy may be suboptimal. 

\subsection{Computational Benefits of ACR with RL}
We compare three RL agents: Q-learning, Q-learning with ACR and Q-learning with Human-Agent Transfer (HAT \cite{taylor2011using}). HAT uses human demonstrations to learn strategies (decision list) from demonstration summaries. We used Q-learning with $\epsilon$-greedy exploration with $N_{ACR} = 50$, $\alpha = 0.25$, $\gamma = 0.99$ and $\epsilon = 0.1$.

We used 5 expert and 5 suboptimal demonstrations separately, to compare the effect of demonstration quality on ACR and HAT. Suboptimal demonstrations refer to where the demonstrator either failed to complete the goal or took a suboptimal path to reach the goal state. In all experiments below, the ACR was built from a single demonstration that exposed the agent to all object-related actions necessary to complete the game. That is, the ACR encodes what the agent \textit{can} do, rather than what the agent \textit{should}. This imposes a more relaxed constraint on the teacher since it is easier to show the ``rules'' rather than the ``strategy'' which requires an expert. Importantly, unlike most existing approaches, the ACR built from expert or suboptimal demonstrations \textbf{do not differ} if the agent learned the same rules from either demonstration. 

Additionally, to evaluate the benefit of entropy-based action selection in RL, the ACR-based approach treats keys and switches as previously unseen objects whose action-categories are unknown and must be simultaneously inferred during the course of the learning process. 

We demonstrate three benefits of using ACR with RL:

\begin{enumerate}
    \item robustness to demonstration quality: we show that ACR has a higher learning rate compared to HAT and Q learning when trained on suboptimal demonstrations
    \item learning from few demonstrations: we show that ACR learns more efficiently than both HAT and Q learning when only a single demonstration is available
    \item improved performance when combining ACR and HAT: we show that best overall performance is achieved when ACR is used to improve the performance of other LfD methods, in this case Human-Agent Transfer.
    
\end{enumerate}

\subsubsection{Effect of Demonstration Quality on ACR:}

\begin{table}[t]
\centering
\includegraphics[width=0.47\textwidth]{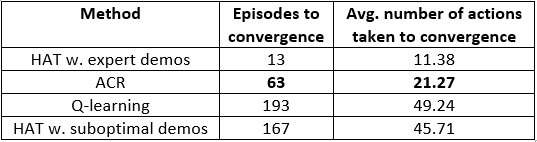}
\captionsetup{justification=centering}
\caption{Comparison of the different approaches based on convergence episode and average number of actions taken by the agent (bold values correspond to ACR)}
\label{fig:RL_table}
\end{table}

\begin{figure}[t]
\centering
\includegraphics[width=0.48\textwidth]{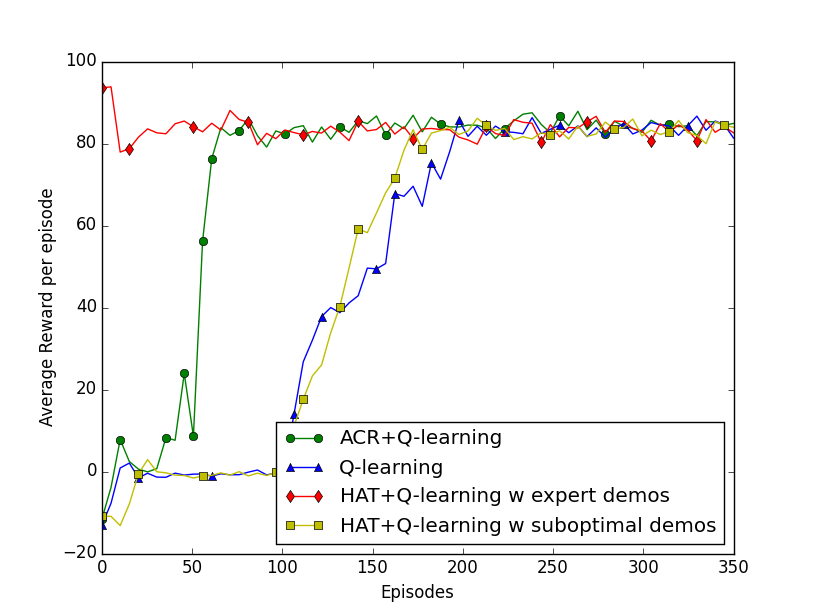}
\captionsetup{justification=centering}
\caption{Comparison of Q-learning, HAT + Q-learning with 5 expert, 5 suboptimal demonstrations and ACR + Q-learning}
\label{fig:avg_R}
\end{figure}

As shown in \ref{fig:avg_R}, ACR performs much better than Q-learning approach and HAT trained on suboptimal demonstrations. ACR also minimizes the number of attempted actions as shown in Table \ref{fig:RL_table}. However, it does not perform better than HAT that uses expert demonstrations. This is because, with enough expert demonstrations, the information contained within ACR can be implicitly learned in the form of rules. Since ACR does not fully leverage the capabilities of a good teacher it does not outperform HAT that uses multiple expert demonstrations. 

However, HAT is quite sensitive to the optimality of the demonstration. The starting reward for HAT is dependent on the teacher performance. As shown in Figure \ref{fig:avg_R}, the starting reward for HAT trained on expert demonstrations is much higher when compared to HAT trained on suboptimal demonstrations. 

To summarize, in cases involving non-expert users, ACR can leverage the rules of the task in order to improve learning performance over the baseline approaches.

\subsubsection{Effect of Number of Demonstrations on ACR:}

\begin{figure}[t]
\centering
\includegraphics[width=0.48\textwidth]{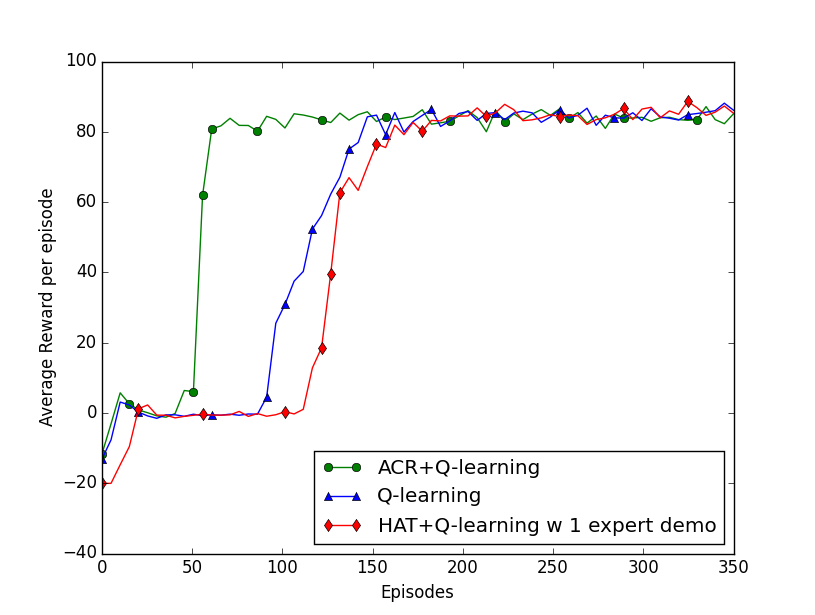}
\captionsetup{justification=centering}
\caption{Comparison of Q-learning, HAT + Q-learning with single expert demonstration and ACR + Q-learning}
\label{fig:avg_R_1_demo}
\end{figure}

Given only a single expert demonstration, HAT fails to accurately summarize the source policy (Figure \ref{fig:avg_R_1_demo}). The building of the decision list in HAT requires more data depending on the complexity of the domain. However, ACR was able to perform better than the baseline approaches with one expert demonstration.  Hence, this makes ACR a feasible approach when there are are not enough demonstrations available to learn a good demonstration policy. 

\begin{figure}[t]
\centering
\includegraphics[width=0.45\textwidth]{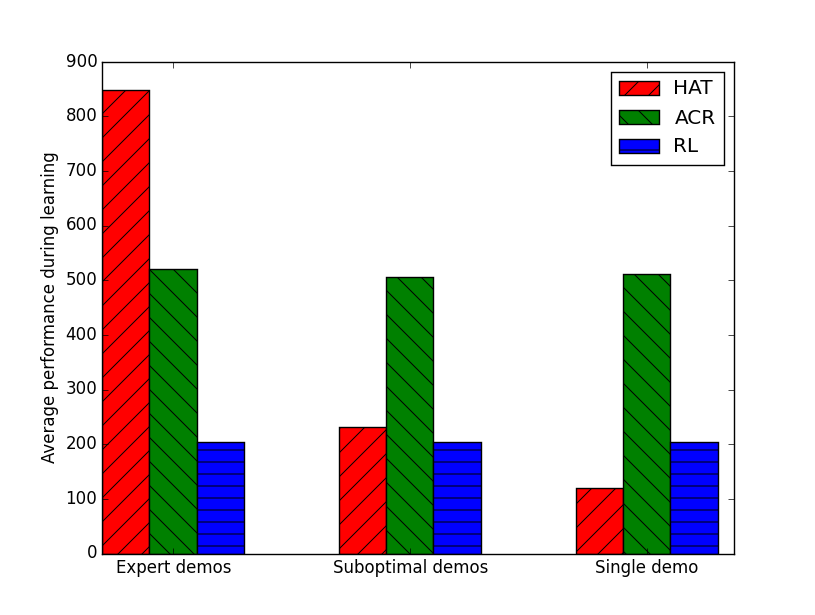}
\captionsetup{justification=centering}
\caption{Summary of the average learning performances based on number and quality of demonstrations for HAT and ACR. Q-learning (RL) also shown for comparison.}
\label{fig:RL_perf}
\end{figure}

Figure \ref{fig:RL_perf} summarizes the effects of number and quality of demonstrations on learning performance of the different approaches. Q-learning is also shown for comparison.

\subsubsection{Combining ACR with HAT}

\begin{figure}[t]
\centering
\includegraphics[width=0.48\textwidth]{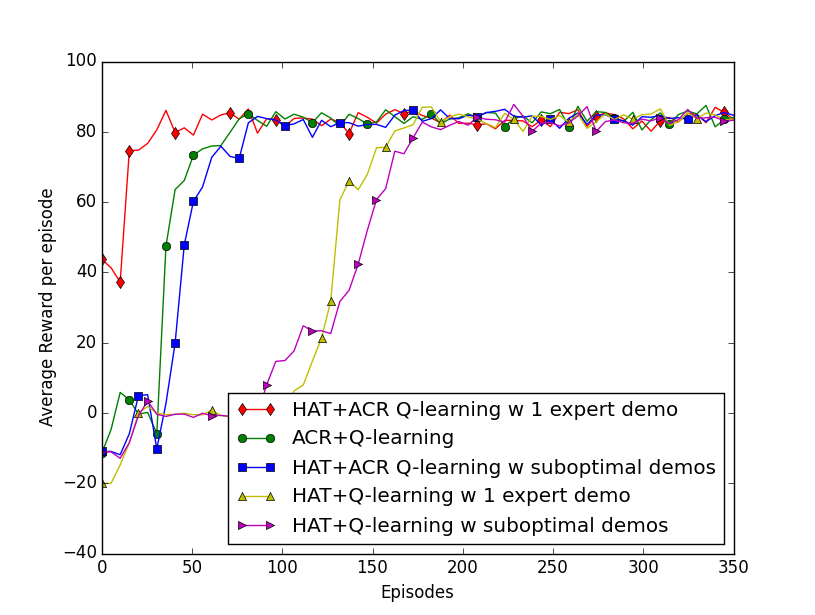}
\captionsetup{justification=centering}
\caption{Performances of HAT + ACR for single expert demonstration and 5 suboptimal demonstrations compared to the baselines that consider the two separately}
\label{fig:avg_R_subopti}
\end{figure}

ACR is an algorithm and domain independent representation; as a result, one of its strengths is that it can be easily combined with complex learning methods, including ones that in themselves influence action selection, such as Human-Agent Transfer. 

HAT consists of 3 steps: demonstration, policy summarization and independent learning. 
Given one or more demonstrations, the teacher's behavior is summarized in the form of a decision list and then used to bootstrap the learning. We utilize the \textit{Extra Action} method from \cite{taylor2011using}, in which the agent executes the action suggested by the decision list for a fixed number of initial episodes before running regular RL.

HAT can bootstrap the learning in states where a good policy is obtained from the demonstrations, while for the ``bad'' states in which the demonstrator's performance was suboptimal, ACR helps accelerate the learning of the optimal policy by reducing the action space. We combine ACR and HAT by verifying that the action suggested by the decision list conforms with the retrieved action category. It does so, by making two checks:
\begin{enumerate}
    \item In states with objects: action selection is restricted to the union of all non-object actions (e.g. movement) and object-related actions within ACR, ensuring that the agent does not try an incorrect action on the object (e.g. ``pick'' button). 
    \item In states without objects: action selection is restricted to non-object actions.
\end{enumerate}

As in the \textit{Extra Action} method, the above action selection method is used to bias exploration early in the learning process before continuing to classical RL using $\epsilon$-greedy Q-learning. 

The results of this method are presented in Figure \ref{fig:avg_R_subopti}, showing improved performance when ACR is combined with HAT. Combining ACR with HAT trained on a single expert demonstration improves the learning performance beyond the case where either of the two approaches are considered separately. Hence, by combining ACR with HAT, it is possible to reduce the effect of number of demonstrations and their optimality on HAT, while also allowing ACR to maximally utilize the teacher demonstrations. 

\section{Conclusion and Future Work}

To conclude, we presented the Action-Category Representation that allows online categorization of objects to action categories based on action codes. Our results demonstrate some of the key benefits of ACR in terms of reduced action space resulting from the action groupings, computational improvements when used with planning and RL, and reduced demonstration requirements with robustness to demonstration errors. 

While the domains described here are discrete in nature, ACR is also applicable to continuous domains by discretizing the state space into states where interaction with an object is possible/not possible. For instance an object may be interacted with, if the agent is within a certain distance of it. Approaches such as \cite{mugan2008continuous} have discussed discretization of continuous state spaces for RL and in this manner, ACR can also be extended to continuous domains. 

In our future work, we aim to address some of the limitations of our work in its current form.  Since ACR currently models single parameter actions, it limits the applicability of ACR to real-world tasks. We plan to address this by incorporating multi-parameter actions by decomposing them into single-parameter actions \cite{bach2014affordance}.  Additionally, future work will explore the possibility of using ACR as a bias with planning (as opposed to the hard constraint on action selection), by utilizing "Ontology-Repair" \cite{mcneill2007dynamic} to update ACR and improve its flexibility. Finally, we wish to extend the application of ACR to Deep Learning for computational improvements in learning performance. 

\section{Acknowledgement}
This work is supported by NSF IIS 1564080.


\bibliographystyle{named}  
\bibliography{references}  

\end{document}